\documentclass[runningheads]{llncs}
\usepackage[T1]{fontenc}
\usepackage{graphicx}
\usepackage{xcolor}
\usepackage{amsmath}
\usepackage{amsfonts}
\usepackage{enumitem}
\usepackage[textsize=tiny]{todonotes} 
\usepackage[hyphens,spaces]{url}
\usepackage{comment}
\usepackage{listings}

\usepackage{algorithm}
\usepackage{algorithmic}

\newcommand{\triple}{\langle h,r,t \rangle}
\newcommand{\triplec}[3]{\langle #1,#2,#3 \rangle}
\usepackage{multirow}
\usepackage{array, makecell}

\begin{document}

\title{Enhancing PyKEEN\\ with Multiple Negative Sampling Solutions\\ for Knowledge Graph Embedding Models}

\titlerunning{Enhancing PyKEEN with Multiple Negative Sampling Solutions}

\author{
    Claudia d'Amato\inst{1,2}\orcidID{0000-0002-3385-987X} \and 
    Ivan Diliso\inst{1}\orcidID{0009-0007-2942-202X} \and 
    Nicola Fanizzi\inst{1,2}\orcidID{0000-0001-5319-7933} \and
    Zafar Saeed\inst{1}\orcidID{0000-0002-1282-5478}
}

\institute{
    Dipartimento di Informatica -- University of Bari Aldo Moro, Italy\\ 
    \email{i.diliso1@phd.uniba.it}, 
    \email{claudia.damato@uniba.it}, 
    \email{nicola.fanizzi@uniba.it}
    \email{zafar.saeed@uniba.it}\and 
    CILA -- University of Bari Aldo Moro, Italy
}

\maketitle              
\begin{abstract}
\textit{Embedding methods} have become popular due to their scalability on
link prediction and/or triple classification tasks on \textit{Knowledge Graphs}. 
Embedding models are trained relying on both positive and negative samples of triples. 
However, in the absence of negative assertions, 
these must be usually artificially generated using various \textit{negative sampling} strategies, ranging from random corruption to more sophisticated  techniques which have an 
impact on the overall performance. 
Most of the popular 
libraries for knowledge graph embedding, 
support only basic such strategies 
and lack 
advanced solutions. 
To address this gap, 
we deliver an extension 
for the popular KGE framework \textit{PyKEEN} 
that integrates a suite of several advanced negative samplers (including both static and dynamic corruption strategies), within a consistent modular architecture, 
to generate meaningful negative samples, while remaining compatible with existing \textit{PyKEEN}-based workflows and pipelines. The developed extension not only enhances \textit{PyKEEN} itself but also allows for easier and comprehensive development of embedding methods and/or for their customization. 
As a proof of concept, 
we present a comprehensive empirical study of the developed extensions and their  impact on the performance (link prediction tasks) of different embedding methods, 
which also provides useful insights for the design of more effective strategies. 


\end{abstract}

\section{Introduction}
\label{sec:introduction}

\textit{Knowledge Graphs} (KGs) represent data in graph structure as factual statements in the form of triples. They provide a complex data representation solution that is  effectively used for numerous knowledge-intensive applications~\cite{gashteovski2020aligning,li2023survey}.
Despite their usefulness, KGs result 
incomplete because of their inherently distributed nature. 
To address this problem, automated completion tasks have gained great importance, 
which has led to the development of efficient \textit{Knowledge Graph Embedding} (KGE) models, resulting from the application of \textit{Representation Learning} to KGs. 
KGE models encode entities and relations described by a KG through a low-dimensional vector space. 
Training such models relies on a contrastive learning approach, where observed positive triples must be distinguished from artificially generated \textit{negative samples} (NS)~\cite{madushanka2024negative}. 
In these models NS is essential for discriminative purposes. 
However, generating accurate negative samples is known to be a critical and challenging task, as KGs consist exclusively of positive statements (\textit{triples}).
NS is typically performed 
by randomly corrupting observed triples, i.e.\ by replacing either their subject or object with an entity randomly sampled from the KG, on the ground of 
the \textit{local closed-world assumption} (LCWA), which posits that the collection is locally complete~\cite{Nickel16}.
This random method may generate poor (trivial or false) negative samples. 
Therefore, NS requires careful differentiation, as the quality of negative samples greatly affects the performance of downstream tasks such as \textit{link prediction} 
that are performed via KGE models. 
Low quality negatives lead to suboptimal embeddings and degraded performance. 
After the foundational work on learning KGE models~\cite{bordes2013translating} 
many advances have been proposed that leverage NS to generate high-quality negatives, focusing on entity similarity, relational semantics~\cite{kotnis2017analysis,dash2019distributional,alam2020affinity}, network structure~\cite{wang2022leveraging}, type-constrained~\cite{krompass2015type}, probability distribution of entities over relations~\cite{wang2014knowledge}, entity aware~\cite{je2022entity}, and adversarial sampling~\cite{cai2018kbgan,zhang2019nscaching}. 
Current methods attempt to exploit the latent semantics in KGs to generate negative samples, but finding a balance between generating meaningful negative samples and avoiding false ones remains a critical challenge. 
The semantics of KGs should be fully and exclusively exploited to get explicit, correct negative statements.  
In principle, the 
approach to generate correct negative statements should be deeply rooted in the semantics of the schema axioms. 
However, despite its high value and availability, little use is made of schema-level knowledge.

In order to devise and experiment advanced approaches to learning quality embedding models, it appears essential to investigate the interplay between orthogonal components: the representation model and the sampling approach adopted for its fitting. 
In fact, given state-of-the-art models, new approaches can be devised by integrating them with alternative NS solutions, so to gain improved 
effectiveness. 
Most of KGE software libraries 
support only basic NS strategies and lack advanced solutions. 
In addition, ad hoc sampling techniques are tightly coupled to the embedding methods provided.
In order to tackle such issues, targeting \textit{PyKEEN}, one of the most popular libraries for building KGE models, we have designed and implemented a modular extension that provides a seamless way to integrate advanced NS techniques with the many existing (and even future) KGE models available, thus multiplying the number of possible combined solutions that can be comparatively investigated. 
Based on this extension, 
the exploration of 
the impact of NS on existing and novel KGE methods (specified in \textit{PyKEEN}) 
is facilitated. As a proof-of-concept, we 
performed an 
empirical comparative study on NS, analyzing state-of-the-art techniques, and 
several KGE models available in \textit{PyKEEN}.

In the following, after reviewing, in Sect.~\ref{sec:basics}, the
    fundamental principles behind negative sampling, the basic and more advanced NS approaches that we implemented within \textit{PyKEEN}, we proceed with illustrating, in Sect.~\ref{sec:resource-description}, the design and development of provided resource, that is 
    a modular implementation of multiple NS techniques as an extension of the popular \textit{PyKEEN} framework for KGE. For showcasing the utility of the developed \textit{PyKEEN} extesion, in Sect.~\ref{sec:exp}, we present 
    a comprehensive experiment design to compare and evaluate the impact of the developed NS solutions, 
    assessing their performance on link prediction tasks.

\section{Basics on Negative Sampling Methods}\label{sec:basics}

The concept of \textit{negative sampling} was first introduced in the context of probabilistic language modeling~\cite{mikolov2013distributed}. 
The inclusion of negative samples simplifies the task by transforming maximum likelihood estimation into a binary classification problem and learning to discriminate between positive and negative samples. 
Training KGE models using negative sampling involves treating (head) entities as analogous to words and their neighboring (tail) entities as their context~\cite{madushanka2024negative}. 
KGE models are trained by distinguishing between positive and negative instances. 
Accurate negative samples enhance the model capacity of grasping the underlying semantics: the quality of negative samples is crucial for optimizing training effectiveness and the models' performance on KG completion tasks.

Since negative triples are not included in the KG, most approaches leverage some form of \textit{Close-World Assumption} (CWA) and systematically corrupt the positive triples to generate negative samples. By imposing various forms of CWA, most negative sampling methods treat non-occurring triples as negatives \cite{chen2023negative}. 
Following this formulation, given the set $\mathcal{E}$ of the KG entities and a positive triple $p=\triple$, with $h,t \in \mathcal{E}$, $p \in \mathcal{T}$ the set of positive triples, one can generate negative triples by replacing either $h$ or $t$ with any entity $e'$ sampled from $\mathcal{E}$, yielding the negative triples $n_t = \triplec{h}{r}{e'}$ (\textit{tail corruption}) and $ n_h = \triplec{e'}{r}{t}$ (\textit{head corruption}) ensuring that $n_h, n_t \notin \mathcal{T}$. 
A notion that is key to formulating specific negative samplers is the \textit{negative pool}, the set of all available entities that can be used to form a correct negative triple when replacing a head or tail; formally:
\begin{eqnarray*}
\mathcal{P}_{head}(\triple) = &  \left\{ h' \mid h' \in \mathcal{E}, \triplec{h'}{r}{t} \notin \mathcal{T} \right\} \\ 
\mathcal{P}_{tail}(\triple) = &  \left\{ t' \mid t' \in \mathcal{E}, \triplec{h}{r}{t'} \notin \mathcal{T} \right\}
\end{eqnarray*}
Given head and tail negative pools, one can randomly sample entities from these sets, corrupt the relative target and produce the negative triples. 
This formulation of negative pool generation is the key difference between the various implemented negative samplers, as it changes the corruption logic for producing ad hoc negative sets for each triple.

\subsection{Static Corruption}

In static corruption, a negative triple is created by defining a criterion for selecting a subset of candidate entities to be used as a negative pool from which to sample the entities. The selection of the negative pool can follow specific criteria \cite{senaratne2023tric,kotnis2017analysis}, such as probability distribution over entities \cite{kotnis2017analysis}, types, and relations. The main difference between these approaches is the logic used to compute the negative pool.

\subsubsection{Random Sampling.}
This is the most commonly used, efficient, and interpretable negative sampler, which defines the negative pool as the entire entity set and samples entities at random. It follows exactly the same formulation as the general one, where the entire entity set is used as a negative candidate pool.
\begin{equation*}
\mathcal{P}_{head}^{random}(\triple) = \mathcal{P}_{tail}^{random}(\triple) = \mathcal{E} 
\end{equation*}

\subsubsection{Bernoulli Sampling} \cite{bordes2013translating}.
Bernoulli shares structural similarities with random sampling in that the negative pool consists of the full set of entities. 
However, the key difference lies in the sampling probability distribution. 
While random sampling treats all entities uniformly (assigning equal probability to each candidate) Bernoulli sampling introduces an asymmetric corruption strategy. 
Specifically, it adjusts the probability of corrupting the head or tail entity by analyzing how the relation typically connects entities, for example, whether it tends to link one entity to many others (1-to-N) or vice versa (N-to-1).
By analyzing the negative pool we can say that:
\begin{equation*}
\mathcal{P}_{head}^{bernoulli}(\triple) = \mathcal{P}_{tail}^{bernoulli}(\triple) = \mathcal{E} 
\end{equation*}

\subsubsection{Corrupt Sampling}
\cite{socher2013reasoning}. 
The corruption strategy is based on the entities that appear as heads or tails for each relation in the positive instance set, leveraging the relation structural information to define the pool of available candidate negative triples. 
Given a triple $\triple$ we can define the negative pools as:
\begin{eqnarray*}
\mathcal{P}_{head}^{corrupt}(\triple) = &  \left\{ h' \mid h' \in \mathcal{E}, \triplec{h'}{r}{*} \in \mathcal{T} \right\} \\ 
\mathcal{P}_{tail}^{corrupt}(\triple) = &  \left\{ t' \mid  t' \in \mathcal{E}, \triplec{*}{r}{t'} \in \mathcal{T} \right\}
\end{eqnarray*}

\subsubsection{Typed Sampling} 
\cite{krompass2015type}. 
This approach takes advantage of the strongly typed relations and entities present in KGs such as DBpedia~\cite{lehmann2015dbpedia} and YAGO~\cite{pellissier2020yago}. 
Given a triple $\triple$, we define $D_r, R_r$ as the domain and range classes of the relation $r$, assuming the dataset has type information in the form of $\triplec{h}{instanceOf}{c}$, we can define the negative pools for each triple as:  
\begin{eqnarray*}
 \mathcal{P}_{head}^{typed}(\triple) = &  \left\{ h' \mid h' \in \mathcal{E}, \triplec{h'}{\mathit{instanceOf}}{D_r} \in \mathcal{T} \right\} \\ 
\mathcal{P}_{tail}^{typed}(\triple) = &  \left\{ t' \mid t' \in \mathcal{E}, \triplec{t'}{\mathit{instanceOf}}{R_r} \in \mathcal{T} \right\}   
\end{eqnarray*}
Basically, the negative pool for a triple head (tail) corruption is the set of all entities that belong to the  domain (range) of its relation. If domain and range properties are not available, a variant of this implementation is defined. In these cases when corrupting head (tail), we use as a negative pool all the entities belonging to the same class of the head (tail) entity:

\begin{eqnarray*}
 \mathcal{P}_{head}^{typed}(\triple) = &  \left\{ h' \mid h' \in \mathcal{E}, \triplec{h'}{\mathit{instanceOf}}{C}, \triplec{h}{\mathit{instanceOf}}{C} \in \mathcal{T} \right\} \\ 
\mathcal{P}_{tail}^{typed}(\triple) = &  \left\{ t' \mid t' \in \mathcal{E}, \triplec{t'}{\mathit{instanceOf}}{C}, \triplec{t}{\mathit{instanceOf}}{C} \in \mathcal{T} \right\}  
\end{eqnarray*}

\subsubsection{Relational Sampling} 
\cite{kotnis2017analysis}. 
This approach makes the assumption that each head-tail pair participates only in one relation. 
Given a triple $\triple$ the negative pools are computed as:
\begin{eqnarray*}
\mathcal{P}_{head}^{relational}(\triple) = &  \left\{ h' \mid h' \in \mathcal{E}, \triplec{h'}{r'}{t} \in \mathcal{T}, r' \neq r \right\} \\ 
\mathcal{P}_{tail}^{relational}(\triple) = &  \left\{ t' \mid t' \in \mathcal{E},  \triplec{h}{r'}{t'} \in \mathcal{T}, r' \neq r \right\}
\end{eqnarray*}

\subsection{Dynamic Corruption}
In contrast to static sampling strategies where the negative pool can be precomputed based on fixed structural or semantic information, dynamic samplers adopt a different paradigm. 
These methods leverage a pre-trained auxiliary model to guide the selection of informative, harder negative samples. 
Specifically, instead of defining the negative pool through schema-based constraints or relational assumptions, the implemented dynamic samplers leverage the entities and predicted entities vector space representation to find more informative negatives. 

In our implementation, we incorporate two dynamic sampling strategies following the approach proposed in~\cite{kotnis2017analysis}. 
These variants differ in how the auxiliary model's learned embeddings are utilized to construct negative pools, reflecting different assumptions about what constitutes an informative negative. 
Let $n$ be the total number of entities, $d$ the dimensionality of the embedding space, $\mathbf{E} \in \mathbb{R}^{n \times d}$ the entity embeddings learned by the auxiliary model and $\text{KNN}(\vec{e}, \mathbf{E}, k) \to \mathbb{N}$ the function that returns the indices of the top $k$ entities closest to $\vec{e}$ from the embedding $\mathbf{E}$ using a (dis)similarity measure of choice.

\subsubsection{Nearest Neighbor}
\cite{kotnis2017analysis}.
In this approach, we use an auxiliary pre-trained model to produce vector representations of entities. 
Given a triple $\triple$, when corrupting on heads (tails) one selects hard negatives by retrieving the $k$  entities nearest to either the head (tail) vector representation. 
This results in a context-aware negative pool, tailored to the embedding geometry. 
Let $\text{Model}_{\text{AUX}} : \mathbb{N} \to \mathbb{R}^{d}$ be the function that returns the embedding vector for entity $e$ using the auxiliary model learned embeddings then, the negative pools for head and tail corruption are defined as follows:
\begin{eqnarray*}
\mathcal{P}_{head}^{knn}(\triple) & =  \text{KNN}(\text{Model}_{\text{AUX}}(h), \mathbf{E}, k) \\ 
\mathcal{P}_{tail}^{knn}(\triple) & =  \text{KNN}(\text{Model}_{\text{AUX}}(t), \mathbf{E}, k)
\end{eqnarray*}

\subsubsection{Adversarial Sampling} 
\cite{kotnis2017analysis}.
This sampling method follows exactly the same formulation as the previous dynamic sampler, but uses the model predictions in the vector space instead of the entity embeddings.
Let $\text{Model}_{\text{AUX}}(e,r,target) \to \mathbb{R}^{d}$ be the function that returns the embedding vector of the prediction on target $target$, given entity $e$ and relation $r$ (for example, in \textit{TransE}, the prediction on tail would be $h+r$ and the prediction on head would be $t-r$). Then the negative pools for head and tail corruption are defined as follows:
\begin{eqnarray*}
\mathcal{P}_{head}^{adversarial}(\triple) & = \text{KNN}(\text{Model}_{\text{AUX}}(h,r, \text{"head"}), \mathbf{E}, k) \\ 
\mathcal{P}_{tail}^{adversarial}(\triple) & = \text{KNN}(\text{Model}_{\text{AUX}}(t,r  \text{"tail"}), \mathbf{E}, k)
\end{eqnarray*}


\section{Resource Description and Technical Details}\label{sec:resource-description}

\begin{figure}[t]
    \centering
    \includegraphics[width=0.80\linewidth]{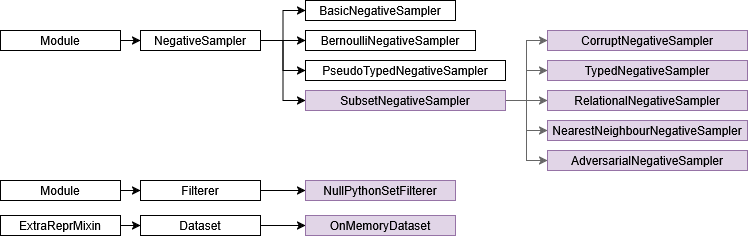}
    \caption{Class inheritance diagram based on \textit{PyKEEN}’s class structure. Custom classes introduced by our extension are highlighted in color.}\label{fig:pykeen_class_schema}
\end{figure}

The proposed resource is an extension to the \textit{PyKEEN} framework, adding a suite of advanced negative sampling strategies for KGE models that are unified under a standardized architecture. The resource is designed to be modular and extensible, and is fully compatible with existing \textit{PyKEEN} workflows. This is achieved by extending the \textit{PyKEEN} abstract classes and adhering to its internal design principles to provide new functionalities within the same standardization. The extension integrates seamlessly with core \textit{PyKEEN} functionalities, such as triple filtering, training, evaluation, and hyperparameter optimisation loops.

\subsection{Core Extensions: Static Sampling}

Static negative sampling refers to strategies in which a predefined pool of candidate entities is used to corrupt triples. Our work builds on this concept by introducing a significant refinement: the negative pool is dynamically adjusted for each triple, enabling context-aware sampling. This means allowing each triple to define its own tailored set of candidates, based on types or structural constraints. The extension is structured around the core class '\texttt{SubsetNegativeSampler}', which inherits from the abstract base class '\texttt{NegativeSampler}' in \textit{PyKEEN}. This provides a generic, reusable, abstract implementation designed to handle the low-level functionalities required to support ad hoc negative pools for each triple. The extension has been developed with extensibility and developer usability in mind. In particular, it isolates the core sampling logic in two abstract methods that custom samplers must implement:

\begin{itemize}
    \item \texttt{generate\_subset()}: This method handles any pre-computation required for sampling, such as filtering by type, frequency, or other criteria. It is designed to avoid redundant computation by preparing intermediate structures beforehand instead of recalculating them for each batch, thereby improving efficiency.
    \item \texttt{strategy\_negative\_pool()}: This method specifies how to calculate the set of actual candidate negatives for a given triple, corruption target and pre-computed subset.
\end{itemize}

By requiring the implementation of only these two methods, the framework abstracts all the 'under-the-hood' complexity, such as batching and tensor operations. Five new negative samplers have been developed following this implementation: "Corrupt", "Typed", and "Relational". These samplers employ various strategies that explore structural and semantic criteria for entity selection, offering an efficient and user-friendly solution within the \textit{PyKEEN} ecosystem. Fig.~\ref{fig:pykeen_class_schema} provides an overview of the implemented samplers in the \textit{PyKEEN} class diagram. Additionally, several functionalities have been implemented in the abstract class, namely:

\begin{itemize}
\item \texttt{choose\_from\_pool()}: A default implementation is provided for randomly sampling entities from the given negative pool. However, developers can override this method and implement custom sampling logic tailored to specific strategies.
\item \texttt{get\_positive\_pool()}: This returns the true positive pool of entities for a given triple and corruption target. This can be used to further refine or filter the negative pool during subset generation.
\item \texttt{average\_pool\_size()}: It computes the average size of the negative pool across all triples.  This metric offers valuable insights into the effectiveness and selectivity of the sampling strategy, enabling more informed decisions about sampler selection.
\item \texttt{integrate}: This parameter specifies whether to supplement the negative pool of the sampler with random entities when the specific corruption criterion is unable to produce a negative pool of the desired size. More details  and an in-depth analysis of this topic is provided in Sect.~\ref{sec:exp}.
\end{itemize}

\subsection{Core Extensions: Dynamic Sampling}

These strategies adhere to the same architectural design and logic as static samplers, and are implemented using the \texttt{SubsetNegativeSampler} base class. Unlike static samplers, however, which pre-compute entity subsets, dynamic samplers need to compute both the subset and the negative pool on the fly at runtime, for each input triple. This design choice is necessary because the negative pool is derived from the predictions of an auxiliary model, which makes pre-computation infeasible. Therefore, model predictions must  be performed dynamically for each triple, based on the current context and model state. To support flexibility and generalization, the dynamic samplers are designed to accept any pre-trained model that implements:\textit{PyKEEN}’s \texttt{ERModel} abstract class. This decouples the auxiliary model from the core sampling logic, enabling developers to define their own training regimes and scoring strategies. The negative sampler requires two inputs:
\begin{itemize}
    \item a pretrained \texttt{ERModel} instance to serve as the auxiliary scoring model;
    \item a user-defined prediction function, which specifies how to compute prediction with the defined \texttt{ERModel}, keeping the predicted entity representation in the vector space.
\end{itemize}

This approach offers maximum flexibility and extensibility, allowing users to define their own logic for computing model predictions. Following this standard, two samplers have been implemented, namely "NearestNeighbour" and "Adversarial".

\begin{table}[tb]
\caption{Available negative samplers in SOTA KGE libraries}\label{tab:kge_lib}
\centering\scriptsize
\begin{tabular}{
l @{\hspace{12pt}} 
l } 
\hline
\textbf{Library}    & \textbf{Samplers}  \\ 
\hline
scikit-kge      & Random, Corrupt \\
OpenKE \cite{han2018openke}         & Random \\
Ampligraph \cite{ampligraph}      & Random \\
GraphVite \cite{zhu2019graphvite}      & Random \\
LibKGE \cite{libkge}         & Random, Bernoulli \\
TorchKGE \cite{arm2020torchkge}      & Random, Bernoulli, Corrupt \\
DGL-KE   \cite{DGL-KE}      & Random \\
PyKEEN  \cite{ali2021pykeen}        & Random, Bernoulli, Corrupt \\
\hline
\textbf{Ours}   & \makecell[l]{Random, Bernoulli, Corrupt, Relational,  \\ Typed (with domain and range), Typed (with only entity classes),  \\ NearestNeighbour, Adversarial} \\
\hline
\end{tabular}
\end{table}

\subsection{Comparison with SOTA libraries}

Tab.~\ref{tab:kge_lib}  presents a detailed comparison of negative sampling support across major open‑source KGE libraries. As shown, most libraries limit negative sampling to uniform or Bernoulli corruption, with little or no built‑in support for more sophisticated strategies. None provide metadata‑driven techniques, and none integrate a standardized fallback mechanism to supplement sparse pools with random samples. Although advanced samplers implementations can be found scattered in separate repositories, our work is the first to offer a fully standardized, modular, negative sampling framework within a widely used library.

\subsection{Documentation, Usability and Compatibility}

Thorough documentation has been provided for the code base to support reuse and encourage community adoption, namely: Python Docstrings following the standard Google style format to ensure clarity and consistency for developers, dedicated READMEs with usage instructions and configuration options, easy-to-use Bash scripts and examples, and consistent code formatting following the 'black' standard, used to promote readability and adherence to Python best practices. 
This structured documentation ensures that the resource is technically robust, accessible, maintainable, and easy to extend. It also aligns with best practices for reusable scientific software.

To further demonstrate  compatibility with the broader \textit{PyKEEN} ecosystem and facilitate user adoption, we provide working examples of each sampler in three usage scenarios: \textit{Training pipelines}, which integrate the samplers into full training scripts using common KGE models; \textit{hyperparameter optimization}, which configure samplers as part of an automated search space;  \textit{standalone use and analysis}, which provides minimal examples that showcase how to instantiate each sampler for inspection, experimentation, and statical analysis.

To demonstrate the ease of integrating our extension into a standard \textit{PyKEEN} pipeline, we provide a minimal example that shows how to register and use a custom negative sampler. The following code snippet shows how a user can use the RelationalNegativeSampler and NullPythonSetFilterer within a \textit{PyKEEN} workflow to integrate the sampler without altering the standard training interface (imports and other details are omitted for brevity):
\begin{small}
\begin{verbatim}
negative_sampler_resolver.register(element=RelationalNegativeSampler)
filterer_resolver.register(element=NullPythonSetFilterer)
pipeline_result = pipeline(
    dataset = "Nations"
    model="TransE",
    negative_sampler="typed"
    negative_sampler_kwargs = dict(
        filtered=True,
        filterer="nullpythonset",
        num_negs_per_pos=params.num_neg_per_pos,
        local_file=/home/me/file.cache
    )
)
\end{verbatim}
\end{small}

\subsection{Additional Extensions}
\subsubsection{Filtering} Handling triples for which no valid negative sample can be generated is a non-trivial issue. For example, this can happen when a strategy relies on type constraints and type information is unavailable for a subset of  entities. To ensure compatibility with \textit{PyKEEN}'s internal computation pipelines, these triples are assigned a entity index value of -1 to serve as a placeholder indicating that they cannot be corrupted by the current strategy. To properly filter out such invalid triples  during training, we implemented the custom filtering class \texttt{NullPythonSetFilterer}, which extends \textit{PyKEEN}'s set-based filter behavior and excludes negatives containing a negative entity index.

\subsubsection{Dataset}
Negative sampling strategies, such as Typed, require additional semantic information. Specifically, they require domain and range properties for relations, and class membership for entities. To enable the use of these strategies within the \textit{PyKEEN} ecosystem, we have developed a custom data loader that extends the functionality of the \textit{PyKEEN} core dataset. This enhanced loader enables users to readily provide and load external semantic metadata alongside standard triple sets. It enables the seamless integration of semantic metadata with standard knowledge graph triples, giving users the ability to supply extra context for negative sampling with minimal setup. The extended loader supports the ingestion of the following external metadata:

\begin{itemize}
    \item Domain and Range: A JSON file specifying the domain and range classes associated with each relation;
    \item Class Membership: A JSON file mapping each entity to its associated semantic classes;
    \item Precomputed IDs Mappings: JSON files that map entities and relations to internal identifiers, ensuring consistency.
\end{itemize}

Additionally, four datasets that adhere to the aforementioned standards have been provided in the repository. Each dataset comes with precomputed ID mapping for entities and relations, as well as triples splits (training, testing and validation) in plain text tab-separated value files. Optional metadata files are also included. This standardized formatting promotes consistent usage and reproducibility across experiments. Specifically,  data and metadata (if available) are provided for \texttt{YAGO4-20}~\cite{pellissier2020yago}, and its subset used in~\cite{barile2024explanation}, \texttt{DBPedia50K} \cite{lehmann2015dbpedia},  \texttt{WN18}~\cite{miller1995wordnet} and \texttt{FB15K}~\cite{bollacker2008freebase}. 



\section{Experimental Design}\label{sec:exp}

This section presents the experimental framework used to validate the proposed extension. Rather than performing an exhaustive benchmarking of KGE models, the objective is to demonstrate the functionality, integration, and practical applicability of the developed components through a proof of concept. Our experiments aim to verify the samplers' seamless compatibility with \textit{PyKEEN}'s training, evaluation and hyperparameter optimization pipelines, and to leverage the new proposed implementation to gain insight into the applicability of negative sampling strategies to a wider range of data.

\subsection{Analysis of the Number of Available Negatives}

The number of negatives generated for each positive instance is a critical parameter in the evaluation of KGE models. In standard random-based approaches, a corrupted entity is sampled from the set of all available entities. However, this assumption poses challenges for methods that employ dynamic or context-specific sampling, since the pool of candidate negatives varies for each triple. In these cases, some triples may lack a sufficient number of valid negative entities.
To mitigate this issue, a strategy commonly adopted in the literature \cite{kotnis2017analysis} involves augmenting the negative pool with randomly sampled entities. While this solution ensures a minimum number of negative samples it introduces a compromise by overriding the intended behavior of the negative sampling strategy. This is particularly problematic for samplers where most triples do not have a negative pool of the desired size. Ultimately, this results in the majority of the training is influenced by random corruption, diminishing the effectiveness of the custom strategy.
To examine this issue, we conducted an empirical analysis of negative pool sizes across four datasets: YAGO4-20, WN18, DBpedia50K and FB15K. We evaluated the ability of each negative sampling method to generate distinct true negatives as the desired number of negatives increased. This highlighted the operational limits of each approach in different datasets. 
In contrast, dynamic samplers behave similarly to random sampling in terms of pool size. However, they differ in that they apply a learned, triple-specific probability distribution to guide the sampling process based on embedding similarity. For this reason, they were excluded from this analysis.

\subsection{Link Prediction Evaluation}

\begin{table}[tb]
\caption{Statistics of the datasets}\label{tab:datasets}
\centering\scriptsize
\begin{tabular}{
    l @{\hspace{12pt}} 
    r @{\hspace{12pt}} 
    r @{\hspace{12pt}} 
    r @{\hspace{12pt}} 
    r @{\hspace{12pt}} 
    r @{\hspace{12pt}} 
    r
}
\hline
\multicolumn{1}{c}{\bf Dataset} & \multicolumn{1}{c}{\bf \#Ent} & \multicolumn{1}{c}{\bf \#Rel} & \multicolumn{1}{c}{\bf Train} & \multicolumn{1}{c}{\bf Valid} & \multicolumn{1}{c}{\bf Test} & \multicolumn{1}{c}{\bf Avg.Degree}\\
\hline
FB15K & 14951 & 1345 & 483142 & 50000         & 59071   & 32.31\\
WN18     & 40943        & 18              & 141442       & 5000          & 5000   & 3.45 \\
\hline
\end{tabular}
\end{table}

\paragraph{Benchmark Datasets.} Our experiments and analysis focus on two datasets: one schema-less WN18~\cite{miller1995wordnet}  and one with additional schema information FB15K~\cite{bordes2013translating}. Their main statistics are reported in Tab.~\ref{tab:datasets}. WN18 is extracted from WordNet\footnote{\url{https://wordnet.princeton.edu/}}, an English-language lexical database where words are interlinked with conceptual semantics. FB15K is a subset of Freebase, an extensive database of general human knowledge organized into tuples. 
These datasets are canonical benchmarks widely adopted in the KGE literature. 
Additionally, their smaller scale wrt to other avialble KGs enables faster and more efficient experimentation for demonstrating 
the functionality of our extension.

\paragraph{Knowledge Graph Embedding Methods.} We have selected twelve KGE methods from three different categories: (1) \textit{translation-based} TransE \cite{bordes2013translating}, TransH, \cite{wang2014knowledge}, TransR,  \cite{lin2015learning} and TransD \cite{ji2015knowledge}, (2) \textit{semantic similarity-based} DistMult \cite{yang2014embedding}, RESCAL \cite{nickel2011three}, ComplEx \cite{trouillon2016complex}, SimplE \cite{kazemi2018simple}, and (3) \textit{geometric-based} RotatE \cite{sun2019rotate}, QuatE \cite{zhang2019quaternion}, BoxE \cite{abboud2020boxe}, HolE \cite{nickel2016holographic}. This selection of models, which are already available in the \textit{PyKEEN} ecosystem, ensures that our negative sampling implementations are tested across a diverse range of approaches. Thereby validating their general applicability and integration with different types of KGE architecture.

\paragraph{Negative Samplers.}  All of the negative samplers introduced in Sect.~\ref{sec:basics} 
have been included in the experimental evaluation, except for the nearest-neighbor-based strategy. This method was excluded due to its significantly higher computational demands, as it requires dense similarity computation for each triple, making it not well suited to the limited scope of this proof-of-concept study. Given our objective, we prioritized strategies that align with our goal of broadly applicable experimentation. To make the results comparable to those of other resources and studies, the sampling procedure was performed using the integration of random negatives when a strategy could not provide a large enough pool. To align our experiments with the proposed experimental design in \cite{kotnis2017analysis}, we used \textit{RESCAL} as the auxiliary model.

\paragraph{Hyper-parameter Optimization.}
We conducted hyperparameter tuning to ensure fair and optimal conditions for all experiments. For each KGE model, and for each configuration of the number of negatives per positive triple, hyperparameters were optimized over 20 trials, with a time limit of 4 hours per trial. All models were trained using mini-batches and an $L_2$ regularization term, with the regularization weight $\lambda$ searched in log scale within the range $[10^{-5}, 10^{-2}]$. Optimization was performed using the \textit{Adam} optimizer, with the learning rate also searched in log scale over the range $[10^{-6}, 10^{-2}]$. A margin-based ranking loss was employed, selecting the margin $\gamma$  from the set ${1, 2, 5, 10}$. The number of training epochs, batch size, and embedding dimensionality were fixed at 100, 500, and 100, respectively. This optimization procedure was repeated independently for each setting in the range of negatives per positive: ${1, 2, 5, 20, 50, 100}$.

\paragraph{Evaluation Protocol.}
Following the standard ranking-based evaluation protocol, we use the hit ratio with a cutoff value of 10 (Hit@10) as an evaluation metric. Hit@K measures the proportion of correct entities among the first $n$ entities. The performance of KGE models was evaluated using filtered criteria to prevent  known triples  from influencing  the ranking list.



\section{Discussion of the Results and Comparative Analysis}

\subsection{Negative Pools Statistics}

\begin{table}[t]
\centering\footnotesize
\caption{Statistics of Negative Sampling Strategies}
\begin{tabular}{
    l @{\hspace{12pt}} 
    l @{\hspace{12pt}} 
    l @{\hspace{12pt}} 
    r @{\hspace{12pt}} 
    r @{\hspace{12pt}} 
    r @{\hspace{12pt}} 
    r @{\hspace{12pt}} 
    r
}
\hline
\textbf{Dataset} & \textbf{Sampling} & \textbf{Avg. Pool} &  $<100$ & $<40$ & $<10$ & $<2$ & \textbf{No Pool} \\
\hline
\multirow{3}{*}{YAGO4-20} 
    & Corrupt     & 4679 & 6.40  & 1.18 & 1.05  & 0.01  & 0.01  \\
    & Typed       & 1125 & 31.38 & 31.38 & 16.36 & 16.35 & 16.01 \\
    & Relational  & 7 & 99.76 & 99.33 & 76.25 & 34.13 & 10.82 \\
\hline

\multirow{3}{*}{DBpedia50}
& Corrupt      & 297 & 44.91 & 26.41 & 7.43 & 2.71  & 1.35  \\
& Typed        & 316 & 69.13 & 54.52 & 37.96 & 34.41 & 21.03 \\
& Relational   & 2 & 99.52 & 99.12 & 97.69 & 89.58  & 38.64 \\
\hline

\multirow{3}{*}{FB15K} 
& Corrupt   & 439  & 45.01 & 30.81 & 20.73 & 10.87 & 3.45  \\
& Typed   & 1295 & 44.50      & 26.12      & 12.17      & 4.45      & 1.49      \\
& Relational  &  39   & 96.27 & 76.27 & 16.67 & 2.65  & 0.56  \\
\hline
\multirow{3}{*}{WN18} 
& Corrupt   & 11367 &   0.78 &	0.70 &	0.00 & 0.00 & 0.00  \\
& Typed   &  --& --      & --      & --      & --      & --      \\
& Relational  &  2  & 99.90 & 99.64 & 96.98 & 61.54 & 5.32  \\
\hline

\end{tabular}
\label{tab:ns_stats}
\end{table}

Tab.~\ref{tab:ns_stats} reports the calculated statistics on the size of negative pools generated by each strategy across four datasets. The percentage values represent the proportion of triples with a negative pool size below a certain threshold, while 'Avg.' Pool column shows the average number of distinct negative entities available per triple (averaged over head and tail corruptions). This analysis is essential for assessing the practical feasibility and scalability of different sampling strategies. As shown in the 'Avg. Pool column, it is evident that all strategies produce significantly smaller candidate sets compared to random sampling. For example, in YAGO4-20, the Corrupt strategy reduces the entity pool from 96,910 to just 4,679 entities -- approximately 4\% of the total. This limitation is further highlighted in the 'Typed' strategy, where missing metadata lead to an even smaller negative pool. Relational approaches perform the worst, with up to 99\% of triples having fewer than 100 different negatives across all datasets. This outcome is motivated by its core assumption that each source-target pair participates in only one relation. However, this assumption is too restrictive for datasets with $1$-to-$N$, $N$-to-$N$, and $N$-to-$1$ relational patterns.

\subsection{Analysis of Link Prediction Results}

\begin{figure}[t]
    \centering    \includegraphics[width=0.75\linewidth]{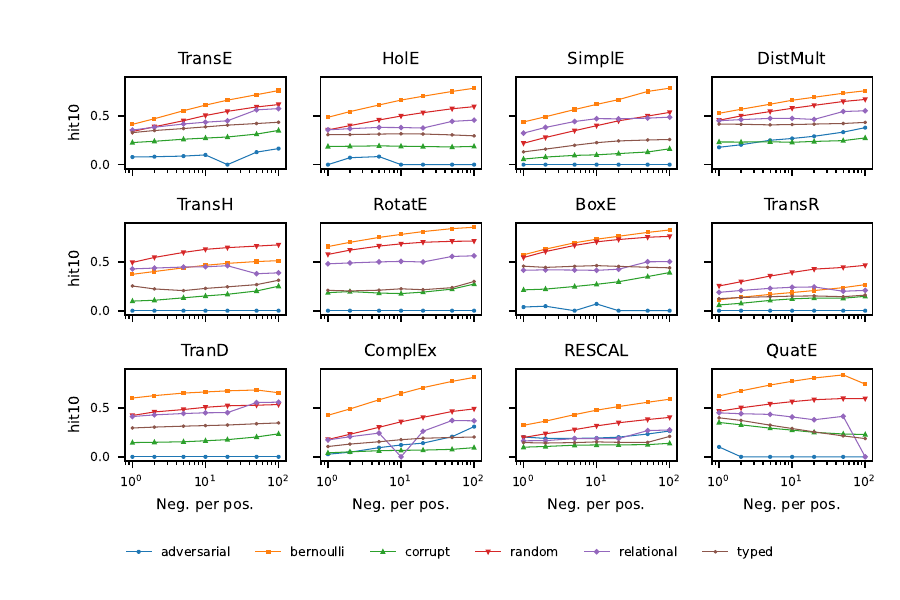}
    \caption{Hits@10 evaluation on FB15K test set}\label{fig:FB15K_result.pdf}
\end{figure}

\begin{figure}[t]
    \centering
    \includegraphics[width=0.75\linewidth]{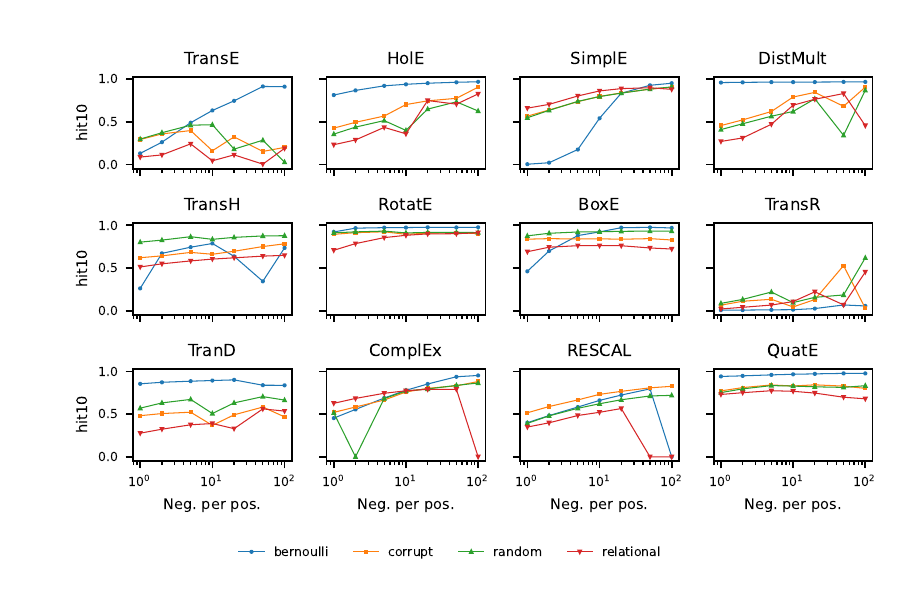}
    \caption{Hits@10 evaluation on WN18 test set}\label{fig:WN18_result.pdf}
\end{figure}

This work presents a proof-of-concept study validating the design and functionality of the proposed extension. Rather than exhaustively benchmarking all samplers, we aim to demonstrate how easily our strategies integrate into the standard \textit{PyKEEN} link prediction pipeline and their empirical viability. To this end, we present the results of link prediction on FB15K and WN18, reporting the Hits@10 ranking metric across multiple values of negative triples generated per positive triple as shown  in Figs.~\ref{fig:FB15K_result.pdf},\ref{fig:WN18_result.pdf}

Results confirm that sampling strategies affect model performance differently depending on dataset structure and sampling design. These effects reflect both the structural properties of the graph  and the design of the sampling algorithm. As well as verifying the correctness and usability of our library, the experiments emphasize that negative sampling choices can have a significant impact on model performance and should therefore be made with an awareness of the characteristics of the dataset and the evaluation objectives.

In FB15K, strategies such as Bernoulli sampling produce strong results across various embedding models, while adversarial remains the least effective. This can be attributed to RESCAL, the auxiliary model used for prediction in this dynamic corruption strategy. According to our evaluation, RESCAL was one of the lowest performing models, which significantly affected the performance of the adversarial sampler.

Taking into account the overall evaluation on negative pool sizes and link prediction metrics, we observe that increasing the number of negative samples per positive instance has a negligible impact on performance, with similar metrics across all models and negative samplers. This can be explained by the use of integration of random negative in the corruption strategy when the negative sampler failed to produce enough candidates. As the analysis shows, many strategies struggle to maintain a meaningful negative pool at higher negative sample sizes. Consequently, as the required number of negatives increases, the majority of training triples become corrupted through random sampling, which reduces the effectiveness of the intended sampling strategy. This results in the sampling behavior becoming increasingly similar to the random corruption scheme as the number of negatives increases. 

These results highlight the limitations of static samplers and suggest practical considerations for their application in research. One key takeaway from this evaluation is the importance of understanding the operational boundaries and dataset compatibility of different negative sampling strategies. It also highlights the need for more in-depth investigations into how negative samplers interact with graph structure and influence model behavior.

\section{Conclusions and Future Works}

This work introduced and validated a modular extension to the PyKEEN framework. The extension has been designed for proving a wide coverage of standardised negative sampler implementations when adopting KGE methods thus filling an important 
gap of the availability of more advanced implemented negative samplers. 
We specifically provided a fully compatible implementation of five negative samplers with static and dynamic corruption strategies.
%
We demonstrated the 
practical utility through a 
range of experiments 
showcasing how the extension can be seamlessly integrated into the training, evaluation and hyperparameter optimisation pipelines of KGE. Furthermore, we presented negative pool statistics for four commonly used datasets, highlighting the operational constraints and behaviour of various sampling strategies in different conditions. 
%
Adhering to PyKEEN architecture and design standards, the proposed extension supports reproducibility, modularity, and ease of experimentation. 
%
By reducing the barriers to developing and evaluating new sampling strategies, our goal is to foster a more unified and efficient research ecosystem.

Future developments will focus on 
additional negative sampling methods and 
a broader experimental evaluation across more diverse datasets. 
An important enhancement lies in performance optimisation, specifically through 
caching strategies and algorithms to accelerate computation. We also foresee the potential for parallel and distributed implementation. 

\paragraph*{Resource Availability Statement:} Source code, pretrained model weights, documentation and datasets (YAGO4-20, DB50K, FB15K, WN18) with their relevant metadata are available from GitHub~\footnote{https://github.com/ivandiliso/refactor-negative-sampler/} and Zenodo~\footnote{https://doi.org/10.5281/zenodo.15413075} (the most up to date information is on GitHub). Readthedocs style documentation can be found on GitHub Pages~\footnote{https://ivandiliso.github.io/refactor-negative-sampler}

\begin{credits}
    \subsubsection{\ackname}
    This work was partially supported by project \emph{FAIR - Future AI Research} (PE00000013), spoke 6 - Symbiotic AI (\url{https://future-ai-research.it/}) under the PNRR MUR program funded by the European Union - NextGenerationEU, and by PRIN project \emph{HypeKG - Hybrid Prediction and Explanation with Knowledge Graphs} (Prot. 2022Y34XNM, CUP H53D23003700006) under the PNRR MUR program funded by the European Union - NextGenerationEU
\end{credits}

\end{document}